\documentclass[letterpaper]{article} 
\usepackage{aaai2027}  

\usepackage[hyphens]{url}  
\usepackage{graphicx} 
\usepackage{natbib}  
\usepackage{caption} 
\usepackage{algorithm}
\usepackage{algorithmic}

\usepackage{newfloat}
\usepackage{listings}
\DeclareCaptionStyle{ruled}{labelfont=normalfont,labelsep=colon,strut=off} 
\floatstyle{ruled}
\newfloat{listing}{tb}{lst}{}
\floatname{listing}{Listing}

\usepackage{booktabs}
\usepackage{amssymb}
\usepackage{amsmath} 
\usepackage{times}
\usepackage{latexsym}
\usepackage{graphicx}
\usepackage{booktabs} 
\usepackage[most]{tcolorbox}
\usepackage{xcolor}
\usepackage{kotex}
\usepackage{xcolor}
\usepackage{enumitem}
\usepackage[T1]{fontenc}
\usepackage[utf8]{inputenc}
\usepackage[table]{xcolor} 
\usepackage{makecell}
\usepackage[most]{tcolorbox}
\nocopyright

\title{Efficient Multilingual Reasoning Transfer via Progressive Code-Switching}
\author{
    Zhijun Wang\textsuperscript{\rm 1}\thanks{Work done during internship at Tongyi Lab.}, Junxiao Liu\textsuperscript{\rm 1}, Hao Zhou\textsuperscript{\rm 1},\\Hao-Ran Wei\textsuperscript{\rm 2}, Baosong Yang\textsuperscript{\rm 2}, Shujian Huang\textsuperscript{\rm 1}\thanks{Corresponding author.}
}
\affiliations{
    \textsuperscript{\rm 1}National Key Laboratory for Novel Software Technology, Nanjing University, China\\  
    \textsuperscript{\rm 2}Tongyi Lab, Alibaba Group\\
    {\{wangzj,junxiaoliu,zhouh\}@smail.nju.edu.cn,\
    {\{funan.whr,yangbaosong.ybs\}@alibaba-inc.com,}\,{huangsj}@nju.edu.cn}\\
}

\newcommand{\method}{PCS}
\begin{document}

\maketitle

\begin{abstract}
Large reasoning models (LRMs) have achieved strong reasoning capabilities in English, yet their performance degrades significantly when required to reason in other languages. A natural solution is to transfer the model's English reasoning ability to target languages. However, existing transfer approaches typically rely on distilled target-language reasoning traces from stronger LRMs or online supervision from external judge models, which are costly and difficult to scale. In this paper, we propose PCS (Progressive Code-Switching), a more efficient transfer framework that requires only lightweight translation — without any stronger model for distillation or judging. PCS first constructs code-switched reasoning traces by translating a subset of English reasoning steps into the target language, and uses them to initialize the model's code-switching ability via supervised fine-tuning. It then applies reinforcement learning with a step-level language consistency curriculum, progressively raising the target-language ratio until the model reasons entirely in the target language. This progressive design provides a smooth transfer path that avoids the instability and performance degradation commonly observed when directly enforcing target-language reasoning. Experiments on multiple benchmarks and five typologically diverse languages show that PCS substantially narrows the performance gap between target-language and English reasoning, yielding more language-consistent reasoning while maintaining competitive accuracy. Code and scripts are freely
available at https://github.com/NJUNLP/PCS
\end{abstract}

\section{Introduction}
Large Reasoning Models (LRMs) such as DeepSeek-R1~\citep{Guo_2025}, Qwen3~\citep{yang2025qwen3technicalreport}, and OpenAI-o1~\citep{openai2024openaio1card} have achieved strong performance on math and code reasoning. Reinforcement Learning with Verified Rewards (RLVR) improves such abilities through scaling, and the long chain-of-thought reasoning steps strengthen problem-solving and enhance interpretability.

Despite these advances, multilingual reasoning—defined as the capability to solve mathematical problems in languages other than English—remains a persistent challenge. When prompted with non-English questions, LRMs often still \textbf{reason in English}~\citep{wang2025polymathevaluatingmathematicalreasoning}. This mismatch between the user's language and the model's reasoning process can undermine comprehensibility and trustworthiness.

Existing approaches to enforce \textbf{input-output} language matching include Prompt-Control~\citep{tam2025languagemattersmultilingualinput}, Prefix-Control~\citep{qi-etal-2025-models}, Supervised Fine-Tuning (SFT) on multilingual traces~\citep{luo2025mmathmultilingualbenchmarkmathematical}, and Reinforcement Learning (RL) with language-consistency rewards~\citep{mistralai2025magistral}. 
However, these methods exhibit distinct practical limitations. Prompt-Control is often unreliable, while prefix-based constraints, though effective for certain languages, frequently compromise accuracy relative to English reasoning. Supervised Fine-Tuning is constrained by the scarcity of multilingual long-reasoning data and tends to introduce repetition artifacts. Furthermore, while RL with language-consistency rewards encourages the model to reason in the question language, performance typically remains inferior to English-based reasoning. 


\begin{figure*}[t]
\centering
\includegraphics[width=0.95\linewidth]{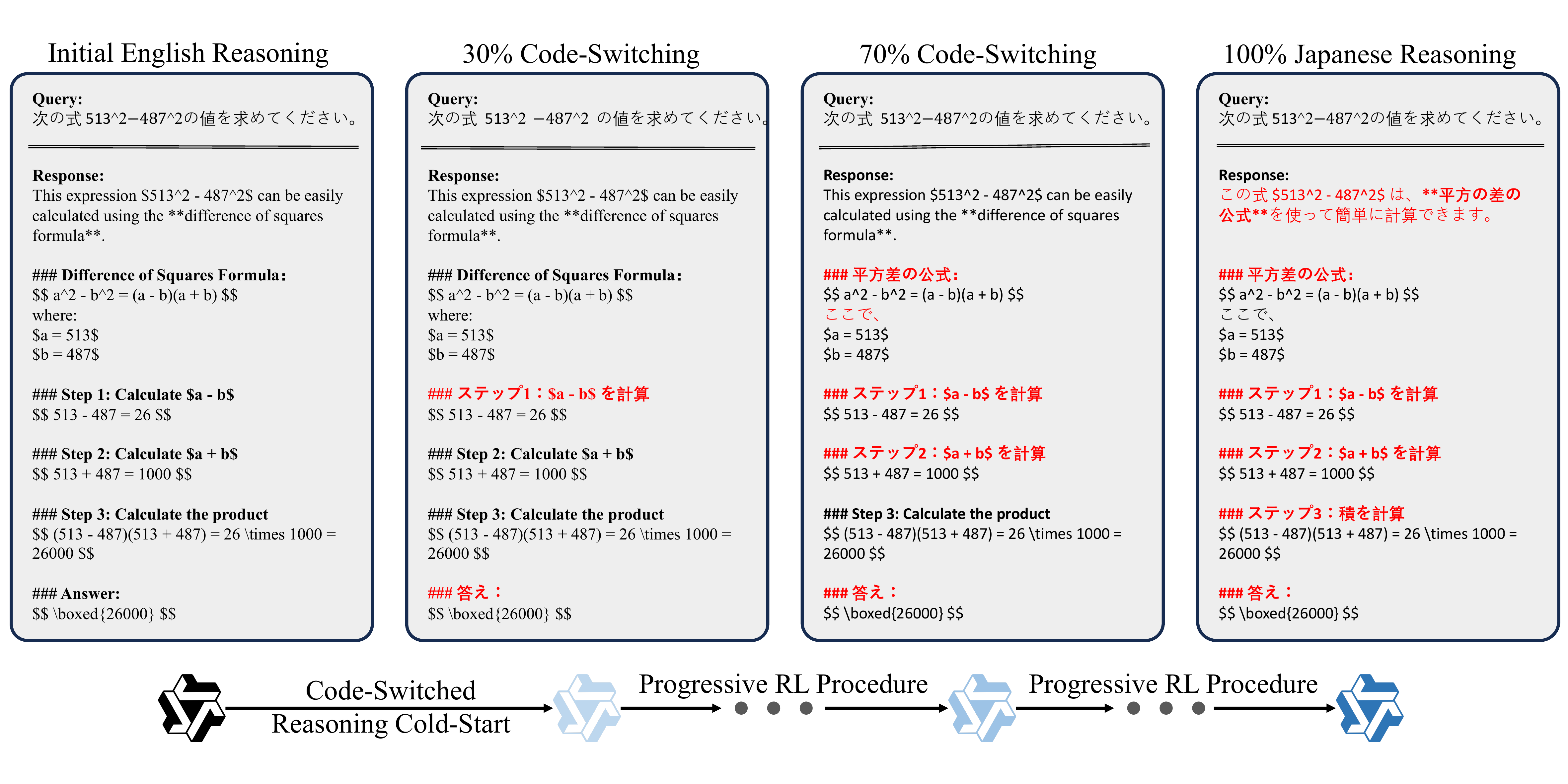}
\caption{The illustration of Progressive Code-Switching. Starting from a cold-start model, \method~progressively increases the target-language reasoning step ratio until the model performs fully target-language reasoning.}
\label{fig:demo}
\end{figure*}

To address these issues, we focus on \textbf{transferring} strong English reasoning ability to target languages. Rather than forcing an abrupt switch to fully target-language reasoning, we propose \method~(\textbf{P}rogressive \textbf{C}ode-\textbf{S}witching), which treats code-switched reasoning as an intermediate stage for cross-lingual transfer. Instead of rewriting all reasoning steps in the target language at once, \method~first converts only a subset of steps, producing mixed English--target-language trajectories that preserve the model’s original reasoning behavior while gradually increasing target-language usage.
As shown in Figure~\ref{fig:demo}, \method~implements this idea with a curriculum over step-level language consistency. Starting from a low target ratio, the model is trained with RLVR to produce correct answers while matching the desired degree of code-switching. As training progresses, the target ratio is gradually increased until the model reaches fully target-language reasoning. This progressive design provides a smoother transfer path and avoids the instability and performance degradation commonly observed under direct target-language supervision or hard language-consistency constraints.

\method~does not require distilled target-language reasoning traces or supervision from a stronger teacher. Experiments across multiple benchmarks and languages show that it consistently improves target-language reasoning consistency while maintaining competitive accuracy, substantially narrowing the gap between reasoning in the target language and in English.

\section{Methodology}
\begin{figure*}[t]
\centering
\includegraphics[width=0.95\linewidth]{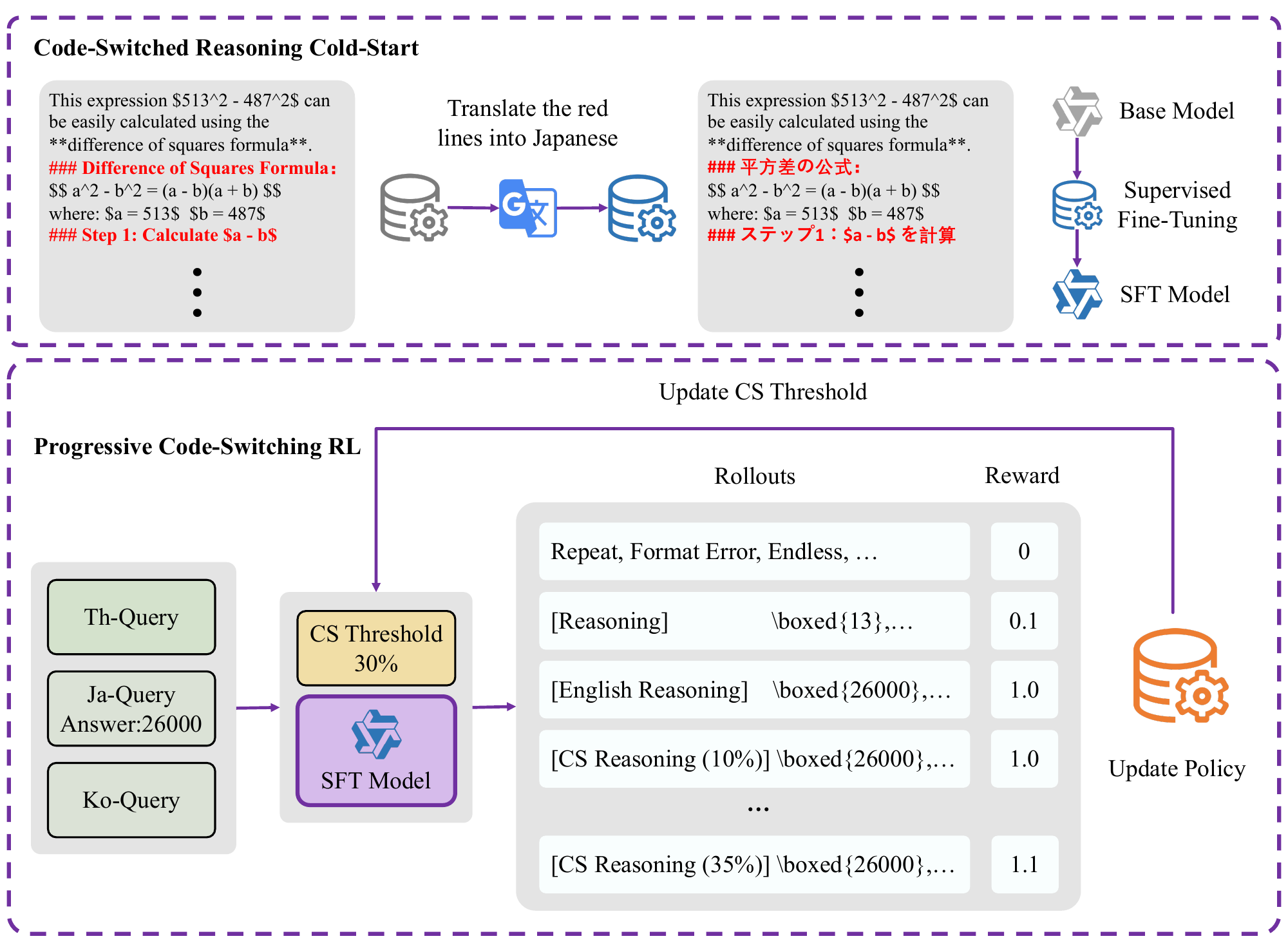}
\caption{    \textbf{The Framework of \method.}}
\label{fig:framework}
\end{figure*}


\subsection{Code-Switched Reasoning}
In this paper, we define \textbf{code-switched reasoning} as an alternating pattern of English and the target language across reasoning steps within a multi-step response, which means some reasoning steps in the original English reasoning traces are expressed using the corresponding target language.
This form of reasoning can encourage the model to exhibit transfer between reasoning modes in the two languages. Although some steps of the high-quality English reasoning are expressed using the target language, the underlying reasoning logic remains the correct and high-quality logic from English; through code-switched reasoning, this logic can gradually be transferred to the target language.

We further define $\mathrm{SLC}(T,L)$, a metric to measure the extent to which the trajectory uses the target language.
Given a reasoning trace $T=(s_1,\dots,s_n)$ and a target language $L$, we assign each step $s_i$
a language tag via a classifier $c(\cdot)\in\{L,\bar{L},\varnothing\}$, corresponding to
target-language text ($L$), non-target-language text ($\bar{L}$), and no natural-language text
($\varnothing$; e.g., steps containing only mathematical expressions).
We define the step-level language consistency as
\begin{equation}
\mathrm{SLC}(T,L)=
\frac{\sum_{i=1}^{n}\mathbf{1}[c(s_i)=L]}
{\sum_{i=1}^{n}\mathbf{1}[c(s_i)\in\{L,\bar{L}\}]}\,,
\end{equation}
where steps with $c(s_i)=\varnothing$ are excluded from the denominator because they are language-agnostic.
If the denominator is zero, we omit this trace when computing batch-level statistics.
We use $\mathrm{SLC}(T,L)$ to quantify the extent to which the trajectory uses the target language,
and aim to push it toward $100\%$ while preserving task performance.

\subsection{Cold-Start}
Since current models often default to English on multilingual math problems, we first perform cold-start SFT on the policy model to enable code-switched reasoning.
We use newline-delimited lines as a practical approximation of reasoning steps, which is lightweight and stable in RL training. Appendix~\ref{potential_hacking} further analyzes the potential problems behind this approximation.
Given each target-language question with an English reasoning trace, we translate a fixed fraction of reasoning steps (30\% in our experiments) into the target language to create a code-switched trace. We then mix these examples with standard English question–answer reasoning data and fine-tune the base model on the combined dataset. This cold-start stage equips the model with an initial ability to produce code-switched reasoning without distilled target-language traces from stronger models.

\begin{table*}[ht]
\centering
\small
\setlength{\tabcolsep}{4pt}
\renewcommand{\arraystretch}{1.3}
\resizebox{\textwidth}{!}{
\begin{tabular}{l|ccc|ccc|ccc|ccc|ccc|c}
\hline
\rowcolor{gray!25}
&\multicolumn{3}{c|}{\textbf{FR}}
&\multicolumn{3}{c|}{\textbf{PT}}
&\multicolumn{3}{c|}{\textbf{JA}}
&\multicolumn{3}{c|}{\textbf{KO}}
&\multicolumn{3}{c|}{\textbf{TH}}
& \textbf{ALL-AVG}\\
\hline
\rowcolor{white}
Methods 
& slc\&acc & acc & slc
& slc\&acc & acc & slc
& slc\&acc & acc & slc
& slc\&acc & acc & slc
& slc\&acc & acc & slc
& slc\&acc \\
\hline
\rowcolor{gray!10}
\multicolumn{17}{c}{MMATH} \\
\hline
Qwen3-4B
&0.0& 28.7& 8.7& 0.0& 29.6& 10.2& 0.0& 23.6& 5.8& 0.0& 24.3& 6.7& 0.0& 24.8& 6.3& 0.0
\\
\rowcolor{white}
Prompt-Control
&0.0& 32.3& 10.9& 0.0& 31.6& 12.3& 0.0& 24.6& 8.1& 0.0& 21.5& 9.3& 0.0& 23.8& 9.1& 0.0
\\
\rowcolor{white}
Prefix-Control
&21.9& 32.4& 88.8& 24.5& 30.8& 95.1& 16.6& 21.3& 96.6& 17.1& 21.8& 91.4& 20.4& 22.3& 94.7& 20.1
\\
\rowcolor{white}
SFT
&2.9& 25.3& 79.2& 3.2& 27.0& 58.8& 7.0& 7.7& 96.6& 9.7& 21.0& 90.2& 11.7& 22.2& 89.6& 6.9
\\
\rowcolor{white}
Naive-RL
&0.0& 33.7& 3.1& 0.0& 32.8& 1.8& 0.0& 32.1& 0.0& 0.0& \textbf{34.6}& 0.1& 0.0& 31.5& 0.2& 0.0
\\
\rowcolor{white}
RLC-RL
&14.6& 29.3& 90.4& 19.6& 25.8& 94.9& 23.2& 27.0& 94.5& 17.1& 24.3& 89.7& 26.6& 27.8& 96.9& 20.2
\\
\rowcolor{white}
M-Thinker
&10.1& \textbf{36.5}& 80.4& 23.7& \textbf{36.4}& 90.2& 29.2& \textbf{35.7}& 87.6& 27.7& 32.4& 92.0& 26.1& \textbf{35.1}& 81.8& 23.4
\\
\rowcolor{gray!8}
PCS
&\textbf{35.1}& 36.0& \textbf{98.1}& \textbf{34.8}& 35.6& \textbf{97.9}& \textbf{31.3}& 32.2& \textbf{97.7}& \textbf{29.5}& 30.1& \textbf{96.9}& \textbf{30.8}& 31.1& \textbf{97.0}& \textbf{32.3}
\\
\hline
\rowcolor{gray!10}
\multicolumn{17}{c}{$\text{MMLU-ProX}_{\text{math}}$} \\
\hline
Qwen3-4B
&0.0& 67.4& 19.3& 0.0& 67.0& 19.3& 0.0& 63.8& 18.2& 0.1& 63.8& 20.6& 0.0& 62.4& 20.3& 0.0
\\
\rowcolor{white}
Prompt-Control
&0.0& 70.4& 21.6& 0.0& 70.0& 22.2& 0.0& 66.8& 22.6& 0.2& 69.0& 25.5& 0.0& 58.6& 23.9& 0.0
\\
\rowcolor{white}
Prefix-Control
&18.8& 34.6& 89.9& 25.4& 38.7& 93.3& 9.9& 12.8& 94.6& 7.0& 10.1& 91.9& 26.5& 33.5& 92.8& 17.5
\\
\rowcolor{white}
SFT
&7.2& 54.5& 80.7& 13.4& 59.8& 71.6& 0.5& 0.6& 96.1& 5.1& 12.4& 88.0& 19.3& 46.4& 90.2& 9.1
\\
\rowcolor{white}
Naive-RL
&0.0& 76.0& 4.8& 0.0& 78.1& 3.0& 0.1& \textbf{76.2}& 0.5& 0.1& \textbf{74.4}& 0.8& 0.0& \textbf{74.5}& 1.6& 0.1
\\
\rowcolor{white}
RLC-RL
&40.4& 72.4& 89.7& 59.2& 73.6& 93.8& 49.3& 67.4& 92.4& 36.1& 65.3& 88.8& 57.9& 66.3& 95.0& 48.6
\\
\rowcolor{white}
M-Thinker
&29.6& \textbf{77.7}& 85.4& 50.1& 76.0& 90.7& 62.1& 75.7& 92.2& 61.0& 73.8& 92.2& 59.9& 71.4& 92.1& 52.5
\\
\rowcolor{gray!8}
PCS
&\textbf{75.1}& 76.9& \textbf{98.1}& \textbf{77.4}& \textbf{78.9}& \textbf{97.9}& \textbf{68.5}& 72.5& \textbf{96.9}& \textbf{65.9}& 70.8& \textbf{95.4}& \textbf{67.9}& 70.6& \textbf{96.1}& \textbf{71.0}
\\
\hline
\end{tabular}
}
\caption{
    Main results on MMATH and $\text{MMLU-ProX}_{\text{math}}$ for Qwen3-4B-Base model.
}
\label{tab:main_result1}
\end{table*}

\subsection{Reward Modeling}
\method~then conducts RL training based on the cold-start model.
Different from conventional response-level language-consistency reward by judging the language of the entire response, we compute the $\mathrm{SLC}(T,L)$ and compare it against a threshold to obtain the language reward. This reward reflects whether the model’s current code-switched reasoning behavior matches the desired level specified by our curriculum.
Given the threshold $\tau$ of $\mathrm{SLC}(T,L)$ at the current optimization step, we first define our reward system:

\begin{itemize}
    \item \textbf{Repetition penalty ($\text{r}_{\text{rep}}$):} We detect repetition in the response to force better reasoning quality.
    $\text{r}_{\text{rep}}=1$ if no repetition, otherwise 0.
    See Appendix~\ref{repeaat} for more details.
    \item \textbf{Format reward ($\text{r}_{\text{fmt}}$):} $\text{r}_{\text{fmt}}=1$ if the output 
    follows the \texttt{<think>...</think>} format, otherwise 0.
    \item \textbf{Accuracy reward ($\text{r}_{\text{acc}}$):} $\text{r}_{\text{acc}}=1$ if the answer 
    is correct, otherwise 0.
    \item \textbf{Step-level language consistency reward ($\text{r}_{\text{SLC}}$):} We apply regular expressions to remove mathematical content. Then use \textit{langdetect}\footnote{\url{https://github.com/Mimino666/langdetect}} to identify the language of each step. $\text{r}_{\text{SLC}}=1$ if $\mathrm{SLC}(T,L) \ge \tau$, otherwise 0. 
\end{itemize}

As demonstrated in Figure~\ref{fig:framework}, we first apply a hard gate \(C=(r_{\mathrm{fmt}}=1 \land r_{\mathrm{rep}}=1)\); if \(C\) is not satisfied, the final reward is set to zero. When \(C\) holds, we assign \(r_{\mathrm{final}}=1\) to correct responses and \(r_{\mathrm{final}}=0.1\) otherwise. For correct responses, we additionally apply a step-level language consistency bonus: \(r_{\mathrm{final}}=1.1\) if \(r_{\mathrm{SLC}}=1\), and \(r_{\mathrm{final}}=1\) if \(r_{\mathrm{SLC}}=0\).

\begin{equation}
r_{\text{final}} =
\begin{cases}
1.1, & \text{if } C \land (r_{\text{acc}} = 1) \land (r_{\text{SLC}} = 1),\\
1,   & \text{if } C \land (r_{\text{acc}} = 1) \land (r_{\text{SLC}} = 0),\\
0.1, & \text{if } C \land (r_{\text{acc}} = 0),\\
0,   & \text{otherwise},
\end{cases}
\end{equation}
\begin{equation}
C = (r_{\text{fmt}} = 1 \land r_{\text{rep}} = 1).
\end{equation}

This reward design not only enforces a high-quality reasoning process but also encourages the model to gradually shift its reasoning language while preserving task performance.

\subsection{Progressive Code-Switching RL}
\label{algo}
Although our initial model acquires preliminary code-switched reasoning ability through cold-start training, its code-switching ratio (i.e., $\mathrm{SLC}(T,L)$) remains low and far from fully target-language reasoning. We then employ reinforcement learning with curriculum learning to increase $\mathrm{SLC}(T,L)$.

Based on our reward modeling, we set $\tau$ at 10\% from the beginning and progressively increase $\tau$ to 95\% during RL training.
We track a batch-level curriculum progress metric, denoted as \(\mathrm{Pass@SLC}(\tau)\). 
It is defined as the ratio of responses achieving \(\mathrm{SLC}(T,L)\ge \tau\) to the total number of responses that satisfy the format and correctness constraints.
Given an adjustment interval $k$ and a margin $\Delta\tau$, we evaluate $\mathrm{Pass@SLC}(\tau)$ every $k$ RL optimization step and update the threshold by
\begin{equation}
\tau \leftarrow 
\begin{cases}
\min(\tau+\Delta\tau,\;0.95), & \text{if } \mathrm{Pass@SLC}(\tau)
\\
&\ge 0.9,\\
\tau, & \text{otherwise}.
\end{cases}
\end{equation}
We set \(k=40\) and \(\Delta\tau=0.1\) in our experiments. When \(\mathrm{Pass@SLC}(\tau)\) reaches 0.9, it suggests that the language-shifting reward becomes saturated (and thus less informative) in the current stage. Increasing \(\tau\) makes the language-consistency reward informative again and corresponds to a harder curriculum, since a higher \(\tau\) requires a larger proportion of reasoning steps in the target language.

We design a progressively increasing \(\tau\) to encourage the model to gradually shift its reasoning language for each step, rather than directly assigning a language reward based on \(\mathrm{SLC}(T,L)\) (e.g., giving a higher reward for a higher \(\mathrm{SLC}(T,L)\)). This is motivated by the fact that transferring reasoning capability from English to the target language via code-switched reasoning is typically a slow adaptation process. An overly aggressive strategy, or one lacking an easy-to-hard curriculum, may cause the model to over-optimize language consistency at the expense of the more important goals of cross-lingual transfer and reasoning performance.

As \(\mathrm{SLC}(T,L)\) increases, the policy distribution deviates more substantially from that of the SFT model, which in turn leads to larger KL divergence and unstable training. In our experiments, an initial KL constraint limits language-switching efficiency, so we mitigate this issue by resetting the reference model to the current policy whenever the KL loss exceeds an empirical threshold of 0.2, and then applying a KL penalty with a coefficient of 0.001.
We apply GRPO to optimize the policy model in our experiments.

%
%
%

\section{Experiments}
\subsection{Experiment Setup}
\begin{table*}[t]
\centering
\small
\setlength{\tabcolsep}{4pt}
\renewcommand{\arraystretch}{1.3}
\resizebox{\textwidth}{!}{
\begin{tabular}{l|ccc|ccc|ccc|ccc|ccc|c|c}
\hline
\rowcolor{gray!26}
&\multicolumn{3}{c|}{\textbf{FR}}
&\multicolumn{3}{c|}{\textbf{PT}}
&\multicolumn{3}{c|}{\textbf{JA}}
&\multicolumn{3}{c|}{\textbf{KO}}
&\multicolumn{3}{c|}{\textbf{TH}}
& \textbf{ALL-AVG}
& \textbf{EN} \\
\hline
\rowcolor{white}
Methods 
& slc\&acc & acc & slc
& slc\&acc & acc & slc
& slc\&acc & acc & slc
& slc\&acc & acc & slc
& slc\&acc & acc & slc
& slc\&acc
& acc \\
\hline
\rowcolor{gray!10}
\multicolumn{18}{c}{MMATH} \\
\hline
Qwen3-8B
&0.0& 27.0& 8.1& 0.0& 25.0& 6.7& 0.1& 23.1& 6.3& 0.0& 22.4& 5.9& 0.0& 22.7& 6.4& 0.0 &24.2
\\
\rowcolor{white}
Prompt-Control
&0.0& 30.1& 10.2& 0.2& 28.1& 9.4& 0.0& 22.0& 7.9& 0.0& 23.6& 8.9& 2.7& 21.8& 17.8& 0.6 & 25.8
\\
\rowcolor{white}
Prefix-Control
&22.9& 30.6& 87.1& 26.7& 30.2& 94.3& 19.2& 19.8& 98.4& 21.2& 22.4& 95.9& 20.9& 21.0& 98.9& 22.2 &24.1
\\
\rowcolor{white}
SFT
&14.1& 32.2& 83.1& 12.8& 34.3& 70.5& 9.8& 9.8& 99.0& 20.8& 23.3& 92.5& 25.3& 25.4& 99.4& 16.6 &25.9
\\
\rowcolor{white}
Naive-RL
&0.1& 41.6& 6.5& 0.2&39.5& 9.0& 0.1& \textbf{40.3}& 3.7& 0.2& \textbf{38.8}& 10.1& 0.1& \textbf{39.9}& 7.1& 0.1 &\textbf{46.1}
\\
\rowcolor{white}
RLC-RL
&30.9& 35.0& 96.5& 32.9& 35.1& 97.0& 30.9& 30.9& \textbf{99.7}& 26.8& 28.1& 98.2& 29.5& 29.8& \textbf{99.8}& 30.2 &39.8
\\
\rowcolor{white}
M-Thinker
&33.2& 41.6& 93.9& 36.4& \textbf{41.2}& 93.7& 35.1& 36.6& 95.7& 31.9& 37.7& 92.7& 33.1& 36.8& 92.1& 34.0 &45.4
\\
\rowcolor{gray!8}
PCS
&\textbf{41.2}& \textbf{42.6}& \textbf{98.0}& \textbf{37.2}& 39.2& \textbf{97.1}& \textbf{36.8}& 38.1& 98.3& \textbf{33.6}& 33.8& \textbf{98.9}& \textbf{36.7}& 37.6& 95.6& \textbf{37.1} &45.0
\\
\hline
\rowcolor{gray!10}
\multicolumn{18}{c}{$\text{MMLU-ProX}_{\text{math}}$} \\
\hline
Qwen3-8B
&0.1& 51.7& 20.6& 0.0& 42.4& 19.6& 0.0& 44.1& 19.2& 0.1& 38.3& 19.8& 0.0& 50.9& 21.8& 0.0 &57.0
\\
\rowcolor{white}
Prompt-Control
&0.0& 68.0& 24.7& 0.2& 67.7& 24.3& 0.0& 65.1& 25.0& 0.2& 67.5& 24.3& 1.3& 58.6& 27.1& 0.4 &55.3
\\
\rowcolor{white}
Prefix-Control
&39.5& 48.4& 90.4& 40.7& 45.8& 94.3& 28.9& 29.1& 98.5& 23.3& 24.0& 98.1& 39.5& 39.8& 99.0& 34.4 &55.8
\\
\rowcolor{white}
SFT
&41.5& 65.6& 90.5& 30.3& 66.3& 80.4& 0.4& 0.4& 99.0& 8.6& 10.0& 95.9& 54.5& 54.7& 99.4& 27.1 &65.9
\\
\rowcolor{white}
Naive-RL
&1.7& 82.6& 16.6& 1.3& \textbf{82.9}& 22.3& 0.7& \textbf{80.3}& 14.1& 2.0& \textbf{79.9}& 30.2& 2.0& \textbf{80.4}& 20.5& 1.5 &\textbf{85.7}
\\
\rowcolor{white}
RLC-RL
&75.2& 80.5& 97.5& 74.5& 80.6& 96.3& 74.9& 75.0& \textbf{99.7}& 70.8& 74.1& 97.8& \textbf{75.4}& 75.6& \textbf{99.8}& 74.2 &83.2
\\
\rowcolor{white}
M-Thinker
&58.4& 79.6& 92.6& 53.7& 80.3& 91.3& 74.6& 76.2& 97.6& 69.8& 78.3& 94.7& 74.0& 77.7& 96.7& 66.1 &83.4
\\
\rowcolor{gray!8}
PCS
&\textbf{79.9}& \textbf{83.1}& \textbf{98.0}& \textbf{78.3}& 81.2& \textbf{97.4}& \textbf{76.8}& 77.2& 99.4& \textbf{75.5}& 77.4& \textbf{98.4}& 75.2& 75.9& 99.5& \textbf{77.1} &85.4
\\
\hline
\end{tabular}
}
\caption{
    Main results on MMATH and $\text{MMLU-ProX}_{\text{math}}$ for Qwen3-8B-Base model.
}
\label{tab:main_result}
\end{table*}
\paragraph{Backbones and Languages}
We adopt Qwen3-4B-Base and Qwen3-8B-Base as the backbone models. We evaluate \method~on a diverse set of languages—French (fr), Portuguese (pt), Japanese (ja), Korean (ko), and Thai (th)—to examine its effectiveness across typologically different languages.

\paragraph{Data}
We design consistent settings for different baselines and ~\method{}. Please refer to Appendix~\ref{setup}.

\paragraph{Evaluation Details}
We evaluate on the math subset of MMLU-ProX~\citep{xuan2025mmluproxmultilingualbenchmarkadvanced} and MMATH~\citep{luo2025mmathmultilingualbenchmarkmathematical}. We use three metrics: \textbf{Step-Level Language Consistency (SLC)}, which measures the proportion of reasoning steps in the target language; \textbf{Accuracy (Acc)}, which measures answer correctness; and \textbf{SLC\&Acc}, the percentage of responses that are both correct and language-consistent (\textbf{SLC} $\geqslant 0.9$), which serves as our primary metric. We use \textbf{SLC} $\geqslant 0.9$ to tolerate a small amount of language-agnostic or stray non-target tokens while still reflecting practically target-language reasoning. 
We report avg@4 on MMATH, which is harder and more stochastic under long reasoning, and avg@1 on $\text{MMLU-ProX}_{\text{math}}$ following the standard zero-shot setting. For MMATH, we macro-average its four subsets (MATH500, CNMO, AIME2024, and AIME2025) to account for their varying difficulty levels.

\paragraph{Baselines}
\begin{itemize}
    \item \textbf{Qwen3-4/8B:} We evaluate the post-trained models in thinking mode under the same context limit.
    
    \item \textbf{Prompt-Control:} \citet{wang2025polymathevaluatingmathematicalreasoning} append language-control instructions at inference time (see Figure~\ref{fig:prompt_control}).

    \item \textbf{Prefix-Control:} Following~\citet{qi-etal-2025-models}, we add ``Okay'' in the question language after ``<think>'' to steer the reasoning language without parameter updates.
    
    \item \textbf{SFT:} Fine-tunes the base model on language-consistent supervised data 
    distilled by DeepSeek-V3.2-Exp.
    
    \item \textbf{Naive-RL:} Optimizes the SFT model only by response correctness.
    
    \item \textbf{RLC-RL:}~\cite{mistralai2025magistral} adds a soft response-level language reward (0.1) to Naive-RL for target-language responses.
    
    \item \textbf{M-Thinker:}~\citet{zhang2026thinknativelyunlockingmultilingual} adds cross-lingual thinking alignment rewards using DeepSeek-V3. We reproduce M-Thinker under our experimental settings.
\end{itemize}

\subsection{Experiment Results}

Tables~\ref{tab:main_result1} and~\ref{tab:main_result} present the main results on MMATH and $\text{MMLU-ProX}_{\text{math}}$. Across both benchmarks and all five target languages (fr/pt/ja/ko/th), \method{} consistently achieves the best overall performance under our primary metric, \textbf{SLC\&Acc}.

\paragraph{\method{} achieves the best balance between correctness and target-language reasoning.}
Compared with the post-trained model and inference-time control methods, \method{} substantially improves \textbf{SLC} while maintaining strong \textbf{Acc}. Prompt-Control provides only weak steering over the reasoning language, whereas Prefix-Control enforces target-language reasoning more aggressively but often hurts accuracy. This suggests that directly constraining the reasoning language at inference time is insufficient for effective transfer of multilingual reasoning. In contrast, \method{} improves language consistency through training-time transfer, leading to much stronger joint performance.

\paragraph{\method{} outperforms both SFT and RL baselines.}
SFT on distilled multilingual traces improves language consistency, but the quality and scale of supervised multilingual reasoning data limit its gains. We also observe that SFT can introduce severe repetition issues in some languages; for example, in Table~\ref{tab:main_result1}, Japanese exhibits a substantial performance drop caused by repetition. Naive-RL preserves relatively strong \textbf{Acc} but typically falls back to English reasoning, resulting in very low \textbf{SLC}. RLC-RL improves \textbf{SLC} with response-level language rewards, yet remains clearly behind \method{} on \textbf{SLC\&Acc}. Overall, these results show that neither SFT nor RL with a language consistency reward is sufficient: effective multilingual reasoning transfer requires a gradual shift of the reasoning language.

\paragraph{\method{} enables stable language transfer with less supervision.}
Across all benchmarks, \method{} achieves the highest \textbf{SLC}, reaching roughly 96--98\% on average while preserving competitive accuracy. Compared with M-Thinker, \method{} attains comparable overall performance without relying on distilled target-language reasoning traces from a stronger LRM or supervision from a stronger judge model. In our reproduction, M-Thinker is more vulnerable to reward-hacking under response-level language control, leading to lower \textbf{SLC} despite relatively strong task performance. In addition, M-Thinker requires online reward evaluation by an external judge model, which introduces non-trivial latency into RL training and can reduce overall hardware utilization. While such overhead may be alleviated through a larger judge-serving setup, this comes at the cost of additional computing resources. By contrast, \method{} only uses lightweight language detection, yielding a more resource-efficient transfer process. We analyze these behaviors in more detail in the following sections.

\section{Analysis}
\subsection{Mitigating Reward Hacking of Language Consistency Reward}

\begin{figure}[h]
\centering
\includegraphics[width=\linewidth]{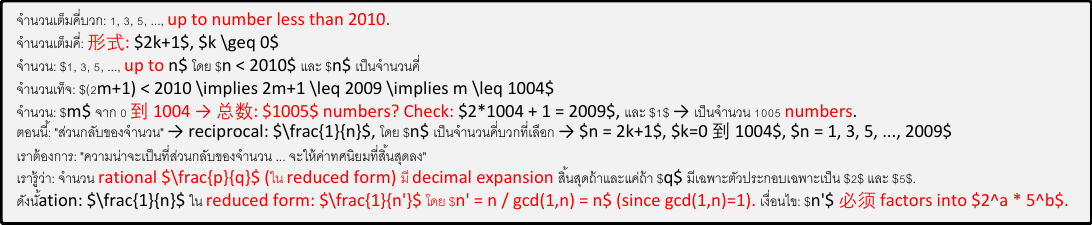}
\caption{Part of M-Thinker's thinking process.}
\label{fig:case-short}
\end{figure}

The RLC-RL and M-Thinker baselines both use response-level language consistency rewards, which reduces language identification to a binary decision based on off-the-shelf language detection tools. In our experiments, this design is largely effective for RLC-RL.
However, the same response-level constraint becomes ineffective for M-Thinker. As shown in Figure~\ref{fig:case-short}, when combined with a cross-lingual thinking alignment (CTA) reward provided by a stronger judge model, M-Thinker often produces low-quality “target-language” reasoning that frequently contains Chinese and English words. This suggests that the policy learns to \textbf{hack} the response-level language constraint: it exploits stronger English/Chinese reasoning patterns to maximize the CTA reward while inserting enough target-language cues to pass a coarse language detector, rather than genuinely target-language reasoning. We provide a detailed case study of this reward-hacking issue in Appendix~\ref{hacking}.

\subsection{Reasoning-Language Transfer Dynamics}
\begin{figure}[h]
\centering
\includegraphics[width=0.9\linewidth]{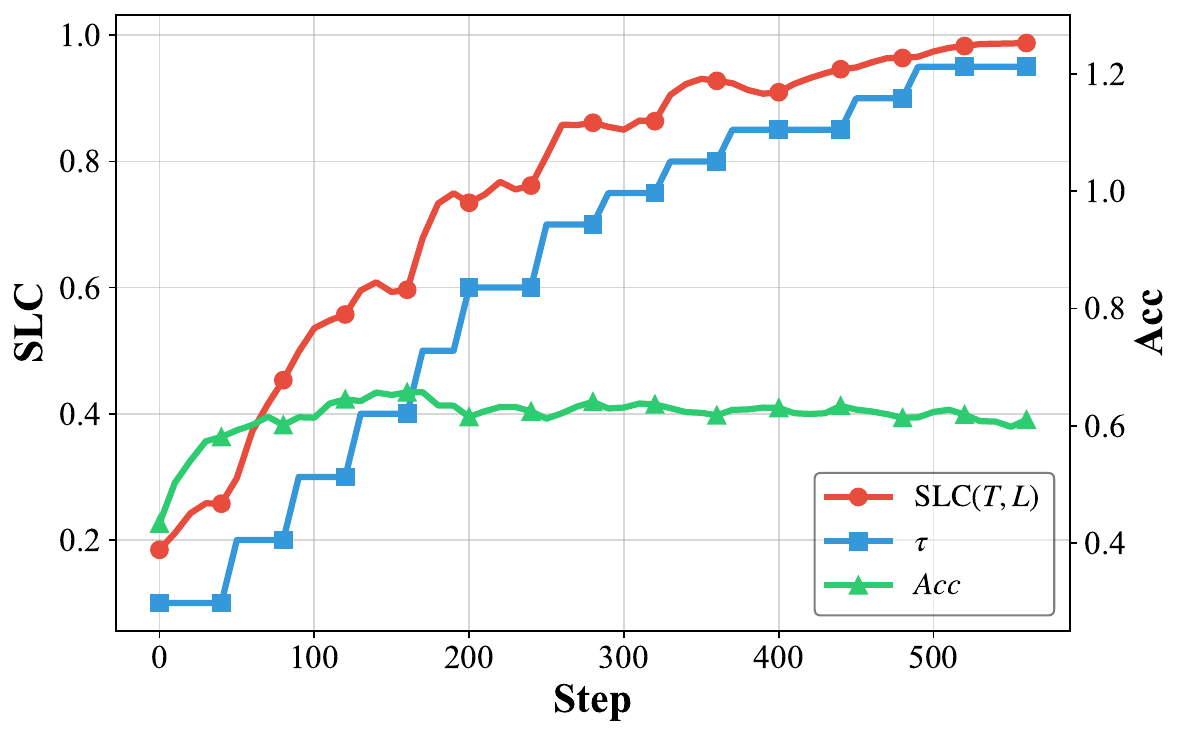}
\caption{Training curves for $\mathrm{SLC}(T,L)$, $\tau$, and Acc of \method.}
\label{fig:slc}
\end{figure}

Figure~\ref{fig:slc} shows the evolution of accuracy, $\tau$, and $\mathrm{SLC}(T,L)$ during training. As $\tau$ increases, $\mathrm{SLC}(T,L)$ steadily rises and eventually approaches 100\%, indicating nearly full target-language reasoning. However, increasing $\mathrm{SLC}(T,L)$ also makes optimization harder, since stronger $\mathrm{SLC}(T,L)$ constraints can conflict with task performance: pushing the model toward fewer English reasoning steps often tends to reduce accuracy. To avoid this trade-off, we assign a higher priority to correctness in the reward design. As a result, accuracy remains relatively stable while $\mathrm{SLC}(T,L)$ improves substantially. Overall, \method{} successfully shifts the reasoning language with only a small performance loss.

\subsection{The Importance of Curriculum}
We introduce a curriculum-learning paradigm into PCS based on $\mathrm{SLC}(T,L)$. To investigate whether this curriculum schedule is necessary, we design an ablation study in which the curriculum is removed and replaced with a reward-shaping strategy that directly and continuously incentivizes higher $\mathrm{SLC}(T,L)$ values.

\begin{equation}
r_{\text{final}} =
\begin{cases}
1 + 0.1 \cdot \mathrm{SLC}(T,L), & \text{if } C \land (r_{\text{acc}} = 1),\\
0.1, & \text{if } C \land (r_{\text{acc}} = 0),\\
0, & \text{otherwise},
\end{cases}
\end{equation}
\begin{equation}
C = (r_{\text{fmt}} = 1 \land r_{\text{rep}} = 1).
\end{equation}

\begin{figure}[h]
\centering
\includegraphics[width=\linewidth]{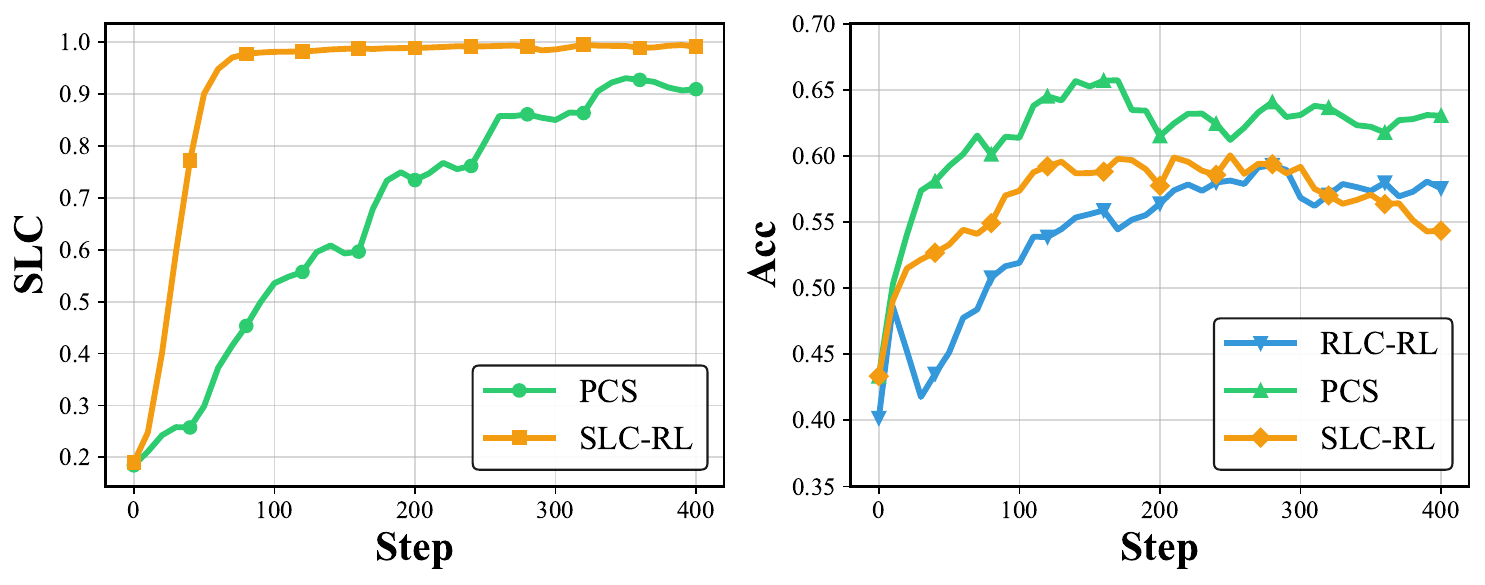}
\caption{The training curves for $\mathrm{SLC}(T,L)$ and Acc for \method, PCS-Dense, and RLC-RL.}
\label{fig:softslc}
\end{figure}

We name this variant \textbf{PCS-Dense} because it uses a dense $\mathrm{SLC}(T,L)$ reward. Figure~\ref{fig:softslc} shows the training curves. Without the curriculum on $\tau$, the policy’s $\mathrm{SLC}(T,L)$ increases rapidly, but the accuracy underperforms \method~ at every step and shows slight improvement over RLC-RL.

We attribute this to the inherently gradual nature of cross-lingual transfer via code-switched reasoning. Because RL is optimized on mini-batches, satisfying the language objective on the current batch does not necessarily generalize to other samples. In contrast, dense rewards push the model to optimize language consistency too aggressively, encouraging premature language switching before the intermediate transfer stage is fully learned and thereby harming reasoning ability.

\subsection{Sensitivity to the SLC Threshold}
\label{bound}
\begin{figure}[h]
\centering
\includegraphics[width=0.9\linewidth]{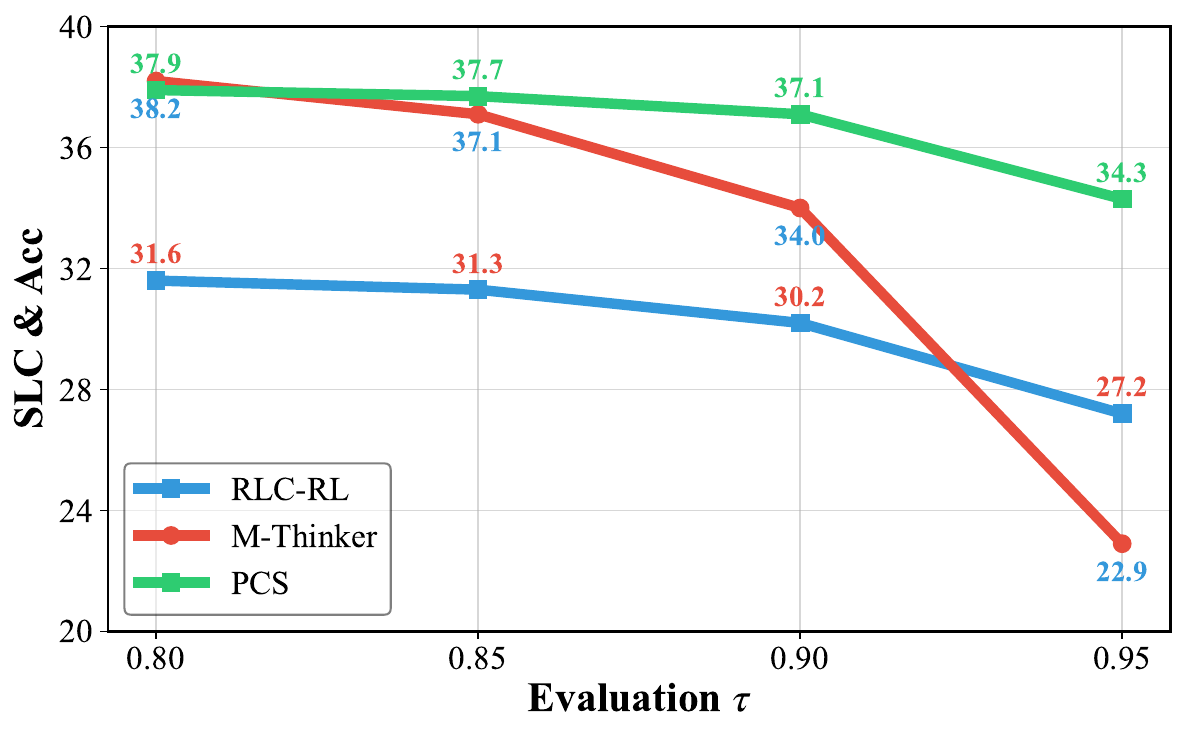}
\caption{The SLC\&Acc results of MMATH across different $\tau$ used in evaluation.}
\label{fig:bound}
\end{figure}

We further examine whether our conclusions depend on the choice of the SLC threshold ($\tau$ used in evaluation) used in the SLC\&Acc metric. Specifically, we evaluate SLC\&Acc under thresholds of 0.80, 0.85, 0.90, and 0.95 for RLC-RL, M-Thinker, and PCS.

As shown in Figure~\ref{fig:bound}, PCS remains substantially more stable as the threshold is tightened. Its SLC\&Acc decreases only mildly, from 37.9 at 0.80 to 34.3 at 0.95. RLC-RL shows a similar but weaker trend, decreasing from 31.6 to 27.2. In contrast, M-Thinker drops much more sharply, from 38.2 at 0.80 to 22.9 at 0.95. Notably, while M-Thinker is competitive under relatively loose thresholds, PCS becomes clearly superior once stricter target-language consistency is required.
These results show that the advantage of PCS is not tied to a particular threshold choice. Instead, PCS is consistently more robust under increasingly strict language-consistency requirements, suggesting that it induces more reliable target-language reasoning rather than merely improving performance near a lenient cutoff. This also supports our use of SLC $\ge$ 0.9 in the main results, which provides a practical balance between enforcing target-language reasoning and tolerating language detection errors.

\section{Related Work}
\subsection{Multilingual Reasoning}
Large reasoning models (LRMs) such as DeepSeek-R1~\citep{Guo_2025} and OpenAI o1~\citep{openai2024openaio1card} achieve strong complex reasoning via RLVR and test-time scaling, yet in multilingual settings, they often exhibit an English bias. DeepSeek-R1 suffers from language mixing and mitigates this with cold-start SFT and language-consistency rewards. However, directly forcing target-language reasoning typically reduces accuracy. Prior work tackles this issue through preference optimization (MAPO~\citep{she2024mapo}), RL with language consistency and cross-lingual thinking alignment (MThinker~\citep{zhang2026thinknativelyunlockingmultilingual}), and multilingual CoT~\citep{lai2024mcot}, training-free structured reasoning~\citep{qi-etal-2025-sot}, and inference-time representation steering~\citep{li2025unlocking}.

\subsection{Code-Switching}
Code-switching, or language alternation, is a linguistic phenomenon where multilingual speakers use multiple languages within a conversation~\citep{poplack1978syntactic}. Code-switching aids multilingual alignment, as demonstrated by \citet{li-etal-2024-prealign}, who use input-only code-switching during pre-training. \citet{yoo2024code} introduces CSCL, a curriculum learning method using synthetic code-switching data to enhance multilingual alignment. 
SynCS~\cite{wang-etal-2025-investigating-scaling} analyzes how natural code-switching enhances LLMs' multilingual capabilities and proposes a more flexible and less expensive code-switching synthesis approach. Overall, code-switching data can facilitate cross-language transfer of models. 
~\citet{son2025pushing} propose Language-Mixed Chain-of-Thought, which permanently anchors reasoning in English while preserving target-language terms at a fixed ratio, treating code-switched reasoning as the desired end state rather than a transitional stage. ~\citet{onyame2026cure} employs code-switching-aware SFT followed by curriculum-informed GRPO, but its curriculum schedules training over language resource tiers with a binary response-level language reward, without explicitly driving the model toward fully target-language reasoning. In contrast, PCS treats code-switching purely as an intermediate bridge and applies curriculum learning directly over the code-switching ratio, progressively raising a step-level language consistency threshold to eliminate code-switching and ultimately achieve complete target-language reasoning. This design requires no distilled traces from stronger teachers or external judge models, and generalizes across typologically diverse languages.  

\section{Conclusion}
We propose \method{} (Progressive Code-Switching), an effective method for transferring English reasoning ability to target languages without requiring distilled target-language reasoning traces from stronger LRMs or supervision from a stronger judge model. By initializing with code-switched reasoning traces and progressively raising a step-level language consistency threshold during RL, \method{} smoothly shifts the model toward fully target-language reasoning while maintaining task performance. Experimental results across five languages demonstrate that \method{} achieves the best Step-Level Language Consistency and overall SLC\&Acc. Furthermore, our analysis confirms that it ensures a stable language transition and improves reasoning quality without reward hacking.

\section*{Limitations}
Our study has several limitations. First, due to limited computational resources, we primarily focus on standard long-reasoning settings and do not explore substantially longer contexts (e.g., 16K or 32K tokens). It remains an open question whether \method{} maintains the same language-shifting stability and capability transfer behavior when the reasoning traces become much longer and require stronger long-context generalization. Second, our experiments are conducted with a relatively small backbone (Qwen3-4B-Base 爱and Qwen3-8B-Base). While \method{} is model-agnostic and does not depend on architecture-specific assumptions, scaling to larger models may change the training dynamics and potentially improve the final multilingual reasoning performance. We leave a systematic study of scaling effects—across both model size and context length—to future work.

\section*{Acknowledgments}
We would like to thank the anonymous reviewers for their insightful comments. Shujian Huang is the corresponding author. This work is supported by National Science Foundation of China (No. 62376116), the Fundamental Research Funds for the Central Universities (No. 2024300507), Fundamental and Interdisciplinary Disciplines Breakthrough Plan of the Ministry of Education of China (No. JYB2025XDXM118).

\bibliography{aaai2027}

\appendix

\section{Detailed Experiment Setup}
\label{setup}
\subsection{Data}
For all training-based methods in our experiments, we first perform a cold-start SFT stage on the base models before RL training. For the baseline methods, the cold-start data consist of distilled target-language reasoning traces, whereas for our method, they consist of code-switched reasoning traces.

For training-based baselines, we largely follow the data recipe of \citet{ji2025amthinkingv1advancingfrontierreasoning}.  
The cold-start SFT data for baselines are trained on 550K samples, including 500K English question–answer reasoning data from \citet{tam2025languagemattersmultilingualinput} and 50K multilingual question–answer reasoning data distilled by DeepSeek-V3.2-Exp (10K per language). The multilingual queries are randomly sampled from the 500K English data and translated to each language by DeepSeek-V3.2-Exp.

The cold-start SFT data for \method~uses the same overall structure, except that the multilingual question–answer portion is replaced with code-switched reasoning data: we randomly select 30\% of the lines in the English reasoning traces and translate them into the target language using TranslateGemma-4B~\citep{finkelstein2026translategemmatechnicalreport}.

For GRPO, we translate all math RL data from \citet{tam2025languagemattersmultilingualinput} into five languages using DeepSeek-V3.2-Exp and use them for all training-based methods.

\subsection{Training Settings}
For SFT on both Qwen3-4B-Base and Qwen3-8B-Base, we train for 2 epochs with a global batch size of 128 and a learning rate of $8\times 10^{-5}$. We pack training samples to a maximum sequence length of 32,768 tokens and set the sequence-parallel size to 4.
For GRPO on both backbones, we train for 600 update steps with a learning rate of $5\times 10^{-6}$, a maximum response length of 4,096 tokens, and a training batch size of 256. Training is fully on-policy. For each prompt, we sample 8 candidate responses during training, using a sampling temperature of 0.9. We use the same temperature (0.9) at test time. We follow the training settings of M-thinker~\citep{zhang2026thinknativelyunlockingmultilingual} on our SFT model and using DeepSeek-V3 as the judge model.
All experiments are conducted on 2x8 H100 GPUs.

\subsection{Repetition Detection}
\label{repeaat}
We employ a multi-level repetition detection mechanism to identify abnormal repetition patterns in model-generated text. Specifically, the detection approach comprises three main components:
\begin{itemize}
    \item \textbf{Line-level exact match detection}: The text is split by line breaks and the frequency of each line's content is counted. A line is identified as repetitive when it repeats beyond a threshold ($\ge$20 times with length $\ge$20 characters, or $\ge$10 times with length $\ge$50 characters).
    \item \textbf{n-gram sequence repetition detection}: Based on suffix arrays and LCP (Longest Common Prefix) algorithms, this method uses divsufsort to construct suffix arrays, computes LCP arrays via the kasai algorithm, and employs sparse tables for efficient range queries to identify n-grams of length $\ge$ 2 with repetition counts $\ge$20.
    \item \textbf{Heuristic detection based on n-gram statistics}: This approach counts the frequency of fixed-length n-grams (default $n=20$). When the highest-frequency n-gram appears $\ge$ 20 times, it undergoes further validation using the suffix array method for precise verification.
\end{itemize}
The detection process executes in order of priority, sequentially performing line-level matching detection, n-gram statistical detection, and suffix array verification. This method effectively identifies various forms of text repetition, including both local vocabulary repetition and global structural repetition.

\section{Multilingual Alignment}
\begin{figure}[h]
\centering
\includegraphics[width=\linewidth]{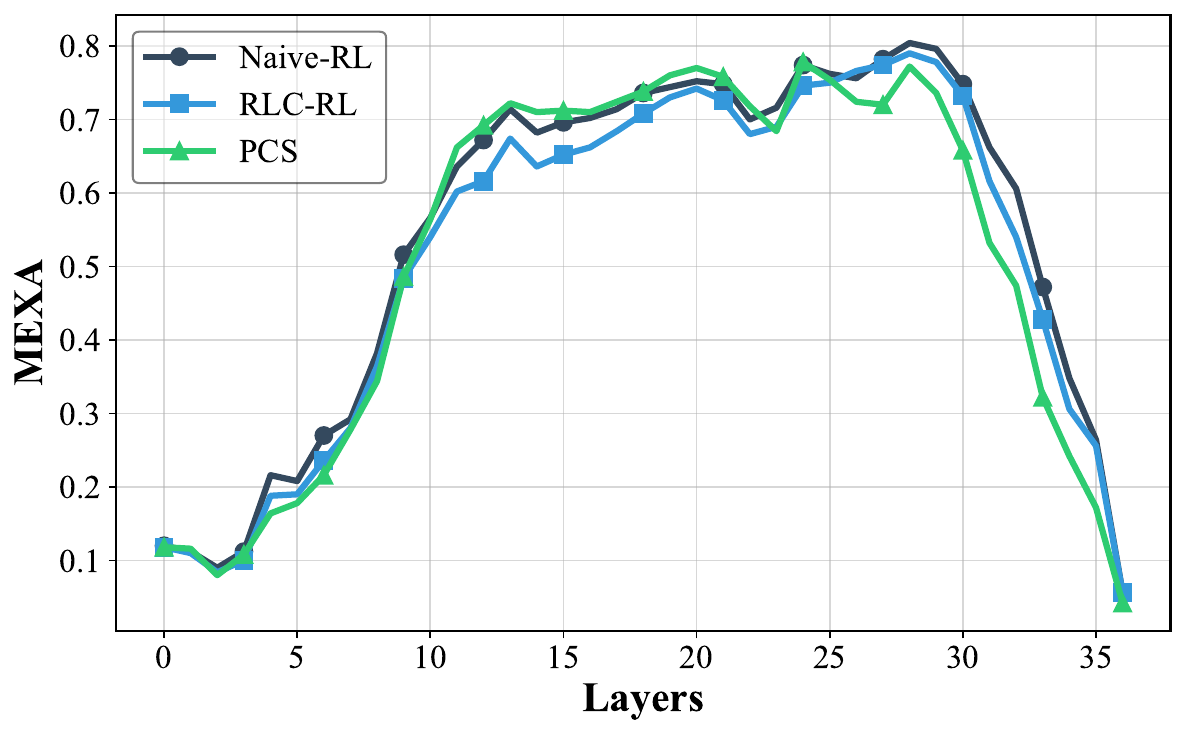}
\caption{The MEXA multilingual alignment score of \method~and baselines across model layers.}
\label{fig:mexa}
\end{figure}
We examine whether \method~aligns its internal representations with English reasoning using the MEXA alignment score~\citep{mexa}. Figure~\ref{fig:mexa} shows the MEXA alignment scores (relative to English) across layers. Prior work~\citep{wendler2024llamasworkenglishlatent} suggests that LLMs typically rely on middle layers for reasoning, lower layers for task understanding, and upper layers for language realization. \method~exhibits a level of alignment comparable to Naive-RL—and higher than SLC-RL—in the middle layers (10–25). Since Naive-RL mainly performs reasoning in English, this result suggests that \method~preserves a stronger alignment with English-style reasoning representations during the reasoning process. In contrast, \method~achieves the lowest alignment scores in the upper layers, indicating larger divergence from English in language realization. Overall, \method~maintains alignment with English reasoning while producing target-language reasoning outputs.

\section{Analysis of Potential Line-Level Reward Hacking}
\label{potential_hacking}
Since PCS measures target-language usage at the \textbf{line} (step) level, one potential failure mode is that the policy may pack many English reasoning tokens into a single long line, while keeping other lines in the target language, thereby satisfying the Step-Level Language Consistency (SLC) constraint while retaining English-style reasoning.

To examine this possibility, we analyze PCS responses on MMATH. For each response, we tokenize every line using Qwen3's tokenizer. We identify the longest line of each response (measured in tokens), and report the mean of this value over the evaluation set. As shown in Table~\ref{tab:length}, PCS indeed tends to produce longer longest-lines than the baselines, suggesting that line length could be a confounding factor for line-level language measurement.

We further test whether the longest line contains substantial non-target-language content. For each response, we apply (i) \texttt{langdetect} to the longest line to obtain a language label, and compute the language consistency (LC) as the percentage of longest lines classified as the target language; and (ii) we use DeepSeek-V3 as a detector to estimate the code-switching ratio within the longest line. The prompt is detailed at Table~\ref{prompt}, and we report the mean of the code-switching label (0/1) over the evaluation set. Table~\ref{tab:hack_analysis} shows that the longest lines are overwhelmingly classified as the target language by \texttt{langdetect} (LC $\ge$ 96.7\% across all languages). Moreover, the estimated intra-line code-switching ratio is very low (0.7--1.6\%), indicating that PCS does not appear to hide large amounts of English reasoning inside the longest line.
Overall, while PCS shows a tendency to generate longer lines, we find no evidence that it exploits this behavior to bypass the line-level SLC constraint. These results support the reliability of our line-level language measurement in practice.

\begin{tcolorbox}[colback=gray!5!white, colframe=gray!75!black, title=Code-Switching Detecting Prompt]
\label{prompt}
\{text\} \\

Please determine whether the language of the given text is consistent, meaning it contains only a single language (excluding mathematical content). \\ \\
Return 0 if it contains only one language, and 1 if it mixes multiple languages. Output exactly 0 or 1 with no additional text or explanations. \\
\end{tcolorbox}

\begin{figure}[h]
\centering
\includegraphics[width=0.9\linewidth]{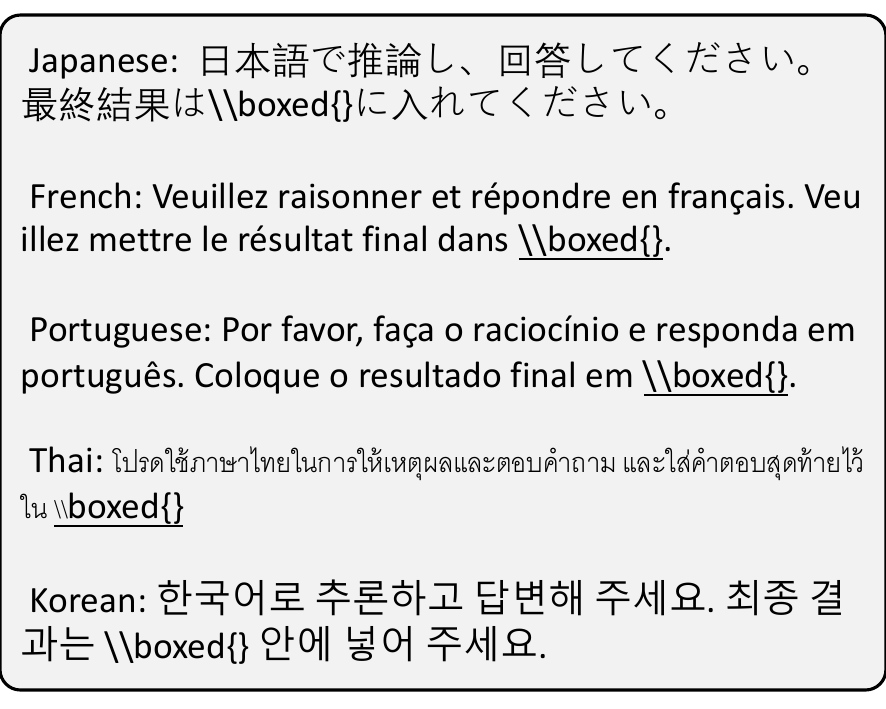}
\caption{Instructions used for Prompt-Control}
\label{fig:prompt_control}
\end{figure}

\begin{table}[]
\centering
\begin{tabular}{lccccc}
\toprule
Method & FR & PT & JA & KO & TH \\
\midrule
Naive-RL &178 &169 &181 &173 &201 \\
RLC-RL &184 &160 &169 &139 &197 \\
M-Thinker &145 &140 &145 &158 &183 \\
PCS &223 &220 &193 &183 &195 \\
\bottomrule
\end{tabular}
\caption{Average token length of the longest line of PCS's response on MMATH.}
\label{tab:length}
\end{table}

\begin{table}[]
\centering
\begin{tabular}{lccccc}
\toprule
Statistics & FR & PT & JA & KO & TH \\
\midrule
LC(\%) &99.1 &98.8 &98.7 &96.7 &99.3 \\
CS Ratio(\%) &0.7 &1.6 &1.2 &1.2 &1.5 \\
\bottomrule
\end{tabular}
\caption{Language Consistency (LC, detect using langdetect tool) and Code-Switching Ratio (CS Ratio, detect using DeepSeek-V3) of the longest line of PCS's response on MMATH.}
\label{tab:hack_analysis}
\end{table}

\section{Reference Update and KL}

\begin{figure}[h]
\centering
\includegraphics[width=\linewidth]{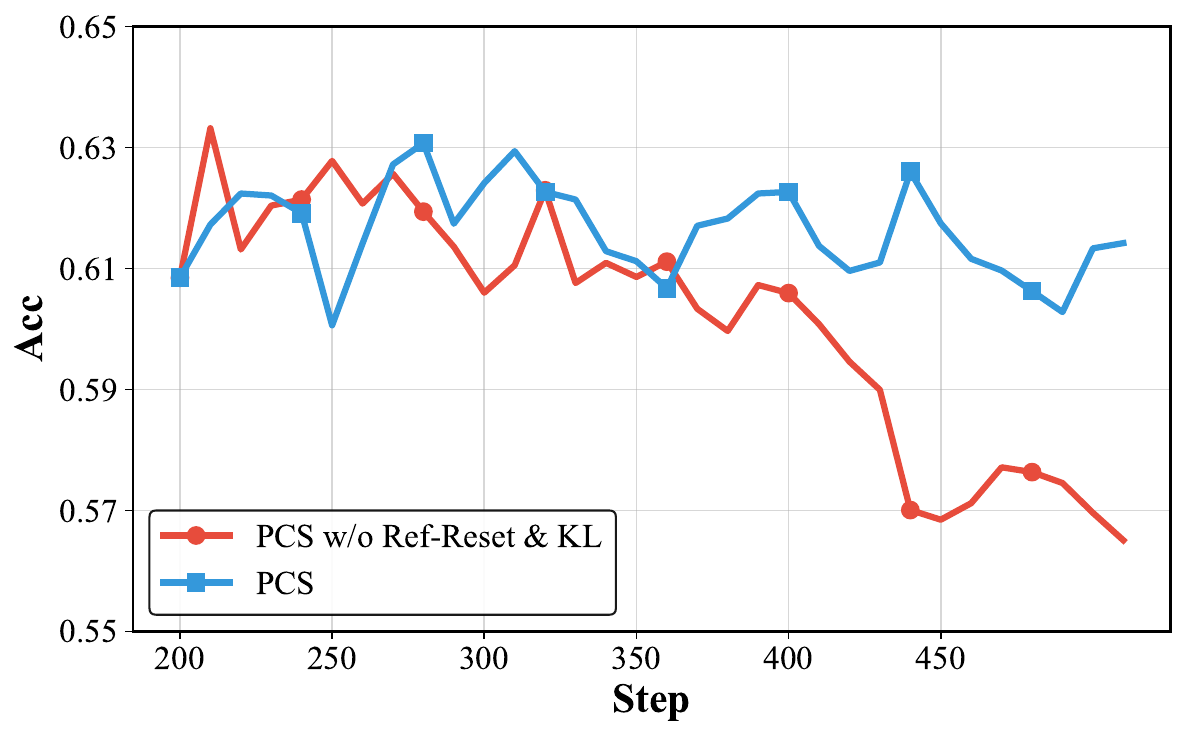}
\caption{The ablation results for \method~ and \method~ without reference-model reset and KL regularization.}
\label{fig:kl}
\end{figure}
As described in Section~\ref{algo}, we reset the reference model and apply KL regularization when the target $\mathrm{SLC}(T,L)$ become large. The reason is that, in the late stage of training, the model is encouraged to produce almost entirely target-language reasoning, making the policy more prone to drifting away from the earlier policy. If left unconstrained, this drift can destabilize RL optimization and hurt final performance. Reference-model reset mitigates this mismatch by updating the optimization anchor to a policy that is better aligned with the current transfer stage, while KL regularization keeps subsequent updates controlled. As shown in Figure~\ref{fig:kl}, removing these components leads to a substantial performance drop, highlighting their role in stabilizing progressive language transfer.

\section{Reward Hacking for Response-Level Language Consistency}
\label{hacking}
\begin{figure}[h]
\centering
\includegraphics[width=0.85\linewidth]{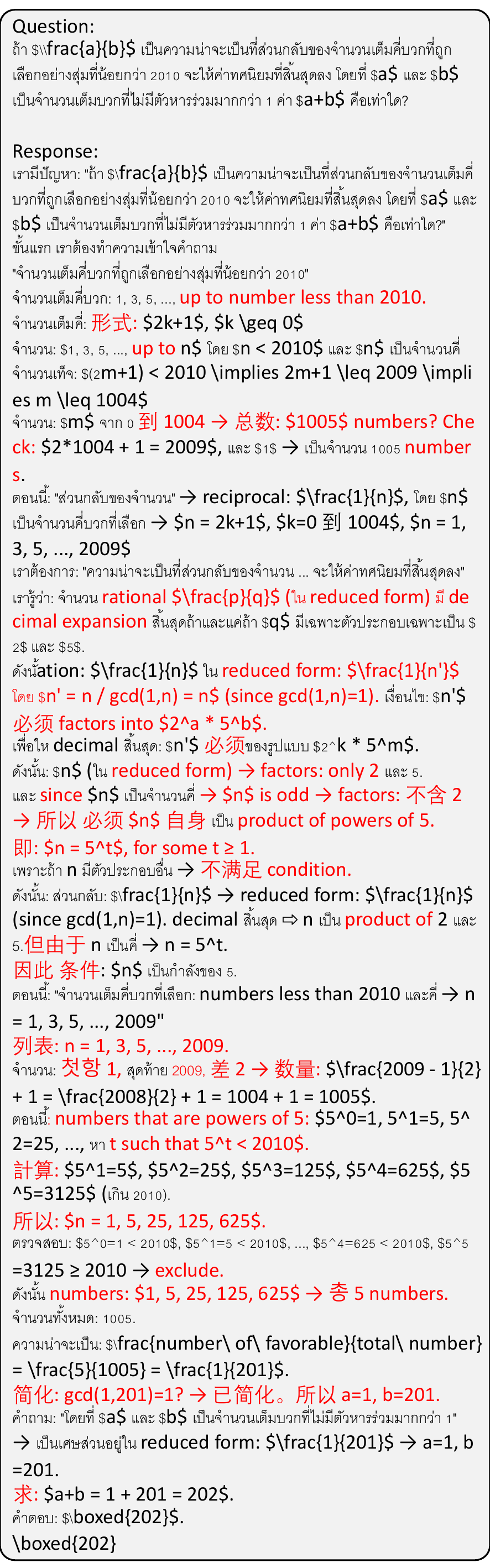}
\caption{Case Study of M-Thinker's model for one sampled Thai math question. The language of this response detected by langdetect is: [th:0.9999999495640537]}
\label{fig:case}
\end{figure}

\begin{figure}[h]
\centering
\includegraphics[width=0.85\linewidth]{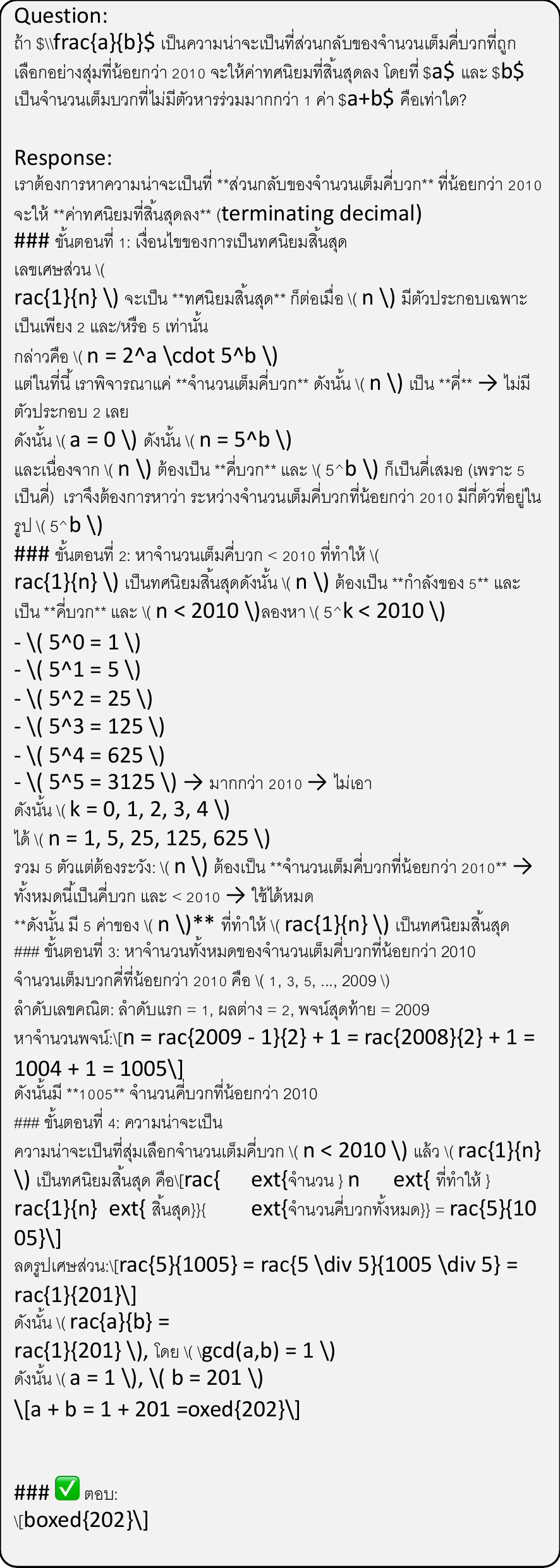}
\caption{Case Study of PCS's model for one sampled Thai math question.}
\label{fig:case1}
\end{figure}

As demonstrated in Figures~\ref{fig:case} and~\ref{fig:case1}, when prompted with a Thai math question, M-Thinker’s response exhibits extensive token-level code-switching between English and Chinese. However, when we apply the \texttt{langdetect} tool to identify the response-level language, it returns ``[th:0.9999999495640537]'', indicating that the response is entirely Thai, which is clearly not the case. Although M-Thinker produces the correct answer, it relies on English and Chinese reasoning abilities and \textbf{hacks} the response-level language detector in a highly unreadable manner.

\end{document}